\documentclass[journal,comsoc]{IEEEtran}
\usepackage[T1]{fontenc}
\usepackage{times}
\usepackage{soul}
\usepackage{url}
\usepackage[hidelinks]{hyperref}
\usepackage[utf8]{inputenc}
\usepackage[small]{caption}
\usepackage{graphicx}
\usepackage{amsmath}
\usepackage{amsthm}
\usepackage{booktabs}
\usepackage{algorithm}
\usepackage{algorithmic}
\usepackage{multirow}
\usepackage{threeparttable}
\usepackage{amsfonts}
\usepackage{float}
\usepackage{url}
\usepackage{bm}
\usepackage{xcolor}
\usepackage{subfigure}
\usepackage{amssymb}

\newcommand{\cmmnt}[1]{}

\ifCLASSINFOpdf
\else
\fi
\usepackage{amsmath}
\usepackage[cmintegrals]{newtxmath}
\begin{document}
%
\title{AnyoneNet: Synchronized Speech and Talking Head Generation for Arbitrary Person}
%
%
%


\author{Xinsheng~Wang,~\IEEEmembership{Student Member, ~IEEE}
    Qicong~Xie,
    Jihua~Zhu,~\IEEEmembership{Member,~IEEE}
    Lei~Xie,~\IEEEmembership{Senior Member,~IEEE}
    Odette~Scharenborg,~\IEEEmembership{Senior Member,~IEEE}
\thanks{Corresponding author: Jihua Zhu and Lei Xie}
\thanks{Xinsheng Wang is with the School of Software Engineering, Xi’an Jiaotong University, Xi’an 710049, China, and also with the School of Computer Science, Northwestern Polytechnical University, Xi’an 710072, China. He also with the Multimedia Computing Group, Delft University of Technology, 2628 CD Delft, The Netherlands. Email: wangxinsheng@stu.xjtu.edu.cn}
\thanks{Qicong  Xie and Lei Xie are with the School of Computer Science, Northwestern Polytechnical University, Xi’an 710072, China. Email: xieqicong@mail.nwpu.edu.cn (Qicong Xie), lxie@nwpu.edu.cn (Lei Xie)}
\thanks{Jihua Zhu is with the School of Software Engineering, Xi’an Jiaotong University, Xi’an 710049, China. Email: zhujh@xjtu.edu.cn}
\thanks{Odette Scharenborg is with the Multimedia Computing Group, Delft University of Technology, 2628 CD Delft, The Netherlands. Email: O.E.Scharenborg@tudelft.nl}
    }

%

%

\markboth{Journal of \LaTeX\ Class Files,~Vol.~14, No.~8, August~2015}%
{Wang \MakeLowercase{\textit{et al.}}: Bare Demo of IEEEtran.cls for IEEE Communications Society Journals}
%



\maketitle


\begin{abstract}
Automatically generating videos in which synthesized speech is synchronized with lip movements in a talking head  has great potential in many human-computer interaction scenarios. In this paper, we present an automatic method to generate synchronized speech and talking-head videos on the basis of text and a single face image of an arbitrary person as input. In contrast to previous text-driven talking head generation methods, which can only synthesize the voice of a specific person, the proposed method is capable of synthesizing speech for any person that are inaccessible in the training stage. Specifically, the proposed method decomposes the generation of synchronized speech and talking head videos into two stages, i.e., a text-to-speech (TTS) stage and a speech-driven talking head generation stage. The proposed TTS module is a face-conditioned multi-speaker TTS model that gets the speaker identity information from face images instead of speech, which allows us to synthesize a personalized voice on the basis of the input face image. To generate the  talking head videos from the face images, a facial landmark-based method that can predict both lip movements and head rotations is proposed. Extensive experiments demonstrate that the proposed method is able to generate synchronized speech and talking head videos for arbitrary persons and non-persons. Synthesized speech shows consistency with the given face regarding to the synthesized voice's timbre and one's appearance in the image, and the proposed landmark-based talking head method outperforms the state-of-the-art landmark-based method on generating natural talking head videos.
\end{abstract}

\begin{IEEEkeywords}
speech synthesis, talking head generation, avatar, facial landmark
\end{IEEEkeywords}

%
\IEEEpeerreviewmaketitle

\section{Introduction}
\label{sc:Introduction}
\IEEEPARstart{A}{utomatically} generating videos in which synthesized speech is synchronised with lip movements in a talking head  has great potential in many human-computer interaction scenarios, e.g., computer games and virtual reality, and in the field of entertainment, e.g., visual dubbing and short video' creation. Intuitively, the synchronized speech and facial animation should not only be dynamically consistent, i.e., the lip and jaw movements should be synchronized to the produced speech, but also perceptively consistent, i.e., the voice should sound like it could be uttered by the person (or non-person) in the video. Otherwise, the generated video would be perceived as unreal and strange. One way to generate talking head videos is to train a model with paired talking head videos and speech, similar to  that in ObamaNet \cite{kumar2017obamanet}. However, a model trained in this fashion can only be used for those persons/faces that are part of  the training process, and such a method thus has very limited generalization. In contrast, in this paper, we present a method that using a still face image of any person and text as input generates a talking head video with a voice that could have been that of the person in the input face image. This method thus works for anyone.

\begin{figure}[tbp]
    \centering
    \includegraphics[width=\linewidth]{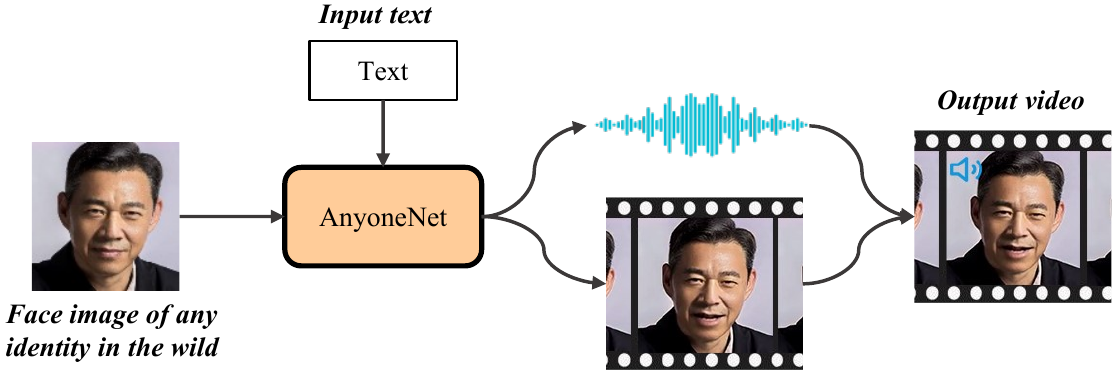}
    \caption{Illustration of generating a talking head video with synchronized speech. The input is text and a still face image, while the output is a talking head video with synchronized speech in which the synthesized voice is in harmony with the person's portrait in the video.}
    \label{fig:Illustration}
\end{figure}

In terms of input (driven) information, the talking head generation methods can be categorized into audio-driven, text-driven, and video-driven \cite{garrido2015vdub,charles2016virtual} methods, i.e., taking audio, text, or video as input to guide the movement of talking heads. Compared to the audio-driven and video-driven methods, the text-driven method is more flexible, as it allows users to create any new content because it is not dependent on an existing corpus or on source videos. Although there are several text-driven methods that directly use textual phonetic labels to predict the visual speech \cite{fan2016deep}, most of the recent text-driven methods \cite{kumar2017obamanet,hati2019text,chae2020text} decompose the text-to-video process into separate text-to-speech (TTS) and speech-to-video processes with a TTS module, i.e., 1) synthesize speech with text as input using the TTS module and 2) perform the audio-driven talking head generation with synthesized speech as input. As a TTS module is indispensable both in the phonetic label-to-video method and text-to-speech-to-video method for building a talking-head video with synchronized audio, in this work, we follow the latter strategy, which allows us to use the intermediate representation of synthesized speech, e.g., spectrograms, to generate the synchronized video.

A high-level overview of our system is illustrated in Fig. \ref{fig:Illustration}. The input face image not only provides identity information for the video generation but also for the TTS system. Specifically, the TTS module tries to synthesize speech with a voice that sounds like it could  have been uttered by the person in the input image. Note however, that unlike research on the reconstruction of face images conditioned on the voice \cite{oh2019speech2face,choi2020inference}, we do not argue that there is a strong relationship between one's portrait and his or her voice. Here, our goal is simply to synthesize a voice that is in harmony with the face in the still image, in order to make the generated voice and the face in the video look natural. 

In terms of the video generation process, both the lip movements and the head movements are predicted using facial landmarks. Different from the state-of-the-art landmark-based method of \cite{zhou2020makelttalk}, in which the head movements are treated as the shift of facial landmarks, here, the head orientation is presented as quaternions, which allows us to predict the head rotations, thus resulting more natural head movements. 

To sum up, the main contribution of this paper is the proposed method that is able to generate voiced talking head video for arbitrary identities. Note that previous work either cannot produce personalized voice for arbitrary persons \cite{chen2019hierarchical,zhou2020makelttalk}, or the voiced talking head video generation can only be used for single person \cite{kumar2017obamanet}. To the best of our knowledge, we are the first to propose this method that can generate synchronized speech and talking head video only with text and a face image of arbitrary person.

The rest of this paper is organized as follows: Section \ref{sc:related works} reviews related work on TTS and audio-driven talking head generation. Section \ref{sc:method} describes the proposed approach. Section \ref{sc:experiments} introduces the databases that are used to train different modules and presents extensive experimental results and evaluation. Section \ref{sc:discussion} discusses the limitations of the proposed method and the ethical considerations are also discussed here. Finally, the paper concludes in Section \ref{sc:conclusion}. Demos of AnyoneNet can be found on the website\footnote{The demos can be found from https://youtu.be/jTb9pyzlHaU}.

\section{Related works}
\label{sc:related works}

\begin{figure*}[t]
    \centering
    \includegraphics[width=\linewidth]{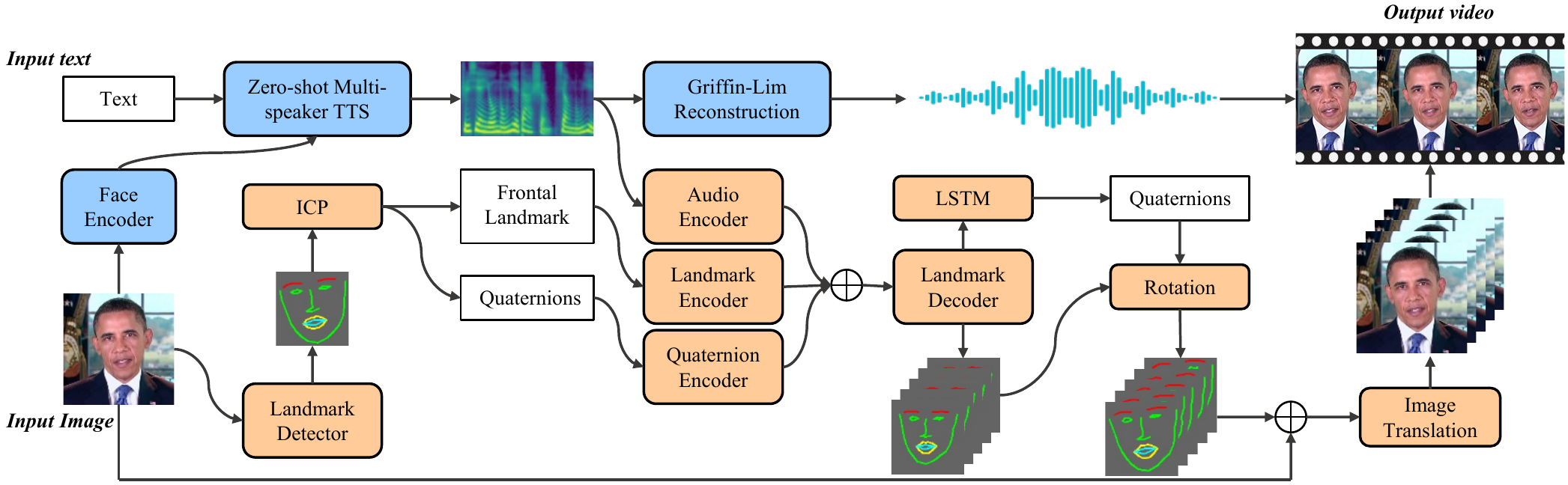}
    \caption{Overall framework of the proposed method. The input is a single still face image and some text. ICP (Iterative Closest Point) is used to register the facial landmarks to a front-facing standard facial template, and resulted rotational parameters are presented as quaternionsß. $\oplus$ indicates concatenation.} 
    \label{fig:Overall_framework}
\end{figure*}

\subsection{Text-to-speech synthesis}
Similar to some recent text-driven talking head generation methods \cite{hati2019text,chae2020text,zhang2021text2video}, our method uses TTS to synthesize the audio track. The goal of a TTS system is to synthesize human-like speech from a natural language text input. 

Most of the recent neural-based TTS methods are performed in two stages. The first stage is to predict low resolution intermediate audio features, typically Mel-spectrograms \cite{shen2018natural,ren2019fastspeech,yu2020durian}, vocoder features \cite{sotelo2017char2wav}, or linguistic features \cite{vanwavenet}, from an input. The second stage is to synthesize the raw waveform audio from the predicted intermediate representation \cite{oord2016wavenet,prenger2019waveglow,kumar2019melgan,kong2020hifi,yang2021multi}. In order to simplify the TTS system in terms of training and deployment, end-to-end TTS models have been proposed \cite{ping2018clarinet,donahue2020end,weiss2021wave}. However, for the talking head generation task, the intermediate representations of the two-stage approach are useful. Therefore, a typical two stage TTS system is adopted in the proposed method, and the intermediate representation Mel-spectrograms are used in the video generation process.  

TTS systems can be categorized into single speaker TTS and multi-speaker TTS systems. The single speaker TTS systems are tailored from a single speaker's voice based on a speech corpus recorded by a single person, e.g., LJspeech \cite{ljspeech17}. In contrast, the multi-speaker TTS systems are able to produce the voices of different speakers. In early research, a multi-speaker TTS model was typically trained as an average voice model using all speakers' data, which was then adapted to an individual speaker \cite{yamagishi2009robust,fan2015multi,yang2016training}. In the recent neural-based methods, conditioning on speaker embeddings has been a popular strategy. Specifically, the speaker representation is commonly extracted by a speaker embedding model and then is used as the conditional attribute in a TTS model \cite{jia2018transfer,nachmani2018fitting,cooper2020zero,casanova2021sc}. For instance, in \cite{jia2018transfer}, the speaker embedding vectors are obtained from a separately trained speaker verification model, and the TTS model Tacotron2 \cite{shen2018natural} conditioned on the speaker embeddings is used for multi-speaker speech synthesis. 

An advantage of the embedding-based multi-speaker TTS is that speaker embeddings can be extracted from any speaker, also speakers who do not exist in the training set, making multi-speaker TTS to be used for any person. To build a talking head generation model that can be used for any person, the embedding-based multi-speaker TTS method is adopted in our TTS module. Different from existing multi-speaker TTS systems, in which the reference speaker embedding is obtained from speech recorded by this speaker, in our method the speaker embedding is based on a person's face image. 

\subsection{Audio-driven talking head generation}

The goal of audio-driven talking head generation  is to create a talking head from a still face image in which lip movements are synchronized with the speech signal. Early methods in this field were usually based on a pre-defined dictionary of visemes, and the model's task was to learn the mappings between the speech signals and the lip articulations \cite{schreer2008real,edwards2016jali}. There are also many efforts from computer graphics to construct 3D models \cite{yu2018bltrcnn,wen2020photorealistic,yi2020audio,thies2020neural,richard2021audio}. However, these 3D model-based methods heavily relay on a person's 3D facial graphic parameters, making them hard to be used for arbitrary persons that are not seen during the training process.

Compared with 3D facial graphic parameters, facial landmarks, i.e., facial key-points, are simpler representations to present the face and mouth shape, which can be easily obtained with recently developed robust and efficient off-the-shelf landmark detectors \cite{king2009dlib,bulat2017far}. A face landmark is to identity the position of a key point on a face, such as the tip of nose and the center of the eye. Each of the points that are detected on the face is called a face landmark. Therefore, facial landmarks can be used to represent the facial-related characteristics, e.g., face shapes, head poses, and mouth shapes, and it is easy to build mapping relation between facial landmarks and the facial expression in a photo. Recently, the facial landmarks have been popular intermediate representations to bridge the gap between the raw audio signal and photo-realistic videos in recent research \cite{suwajanakorn2017synthesizing,hati2019text,fang2019audiovisual,chen2019hierarchical,yu2020multimodal}. However, these methods suffer from several limitations, such as only can be used for the person that used for the training data \cite{suwajanakorn2017synthesizing} and can not be used for arbitrary persons, depending on reference videos to provide pose information \cite{suwajanakorn2017synthesizing,yu2020multimodal}, or no head movement is predicted and can only present static head pose \cite{chen2019hierarchical}. To address these issues, MakeItTalk \cite{zhou2020makelttalk} was developed to disentangle linguistic information and information about the identity of the speaker in the input speech signal. Linguistic information is then used to guide (drive) the lip movements and speaker identity information is to drive facial expressions and head poses. By predicting shifts of landmarks rather than landmarks with specific shapes, MakeItTalk can be easily used for arbitrary identities.    

Most recently, end-to-end models have shown promising results in generating accurate lip movements \cite{chung2017you,vougioukas2018end,zhou2019talking,prajwal2020lip}. However, these methods can only generate a talking head which has a fixed head pose, which limits the naturalness of the generated videos. In order to generate a talking head with more natural head movement, in \cite{zhou2021pose}, a source video is used to provide head pose information which is used to give the predicted talking head the same pose movements. However, to achieve the possibility of altering poses, both the pose information and identity information are represented as embedded vectors, which makes their model cannot be used to an arbitrary person that was not accessible during the training process.

With the goal to build talking head videos with natural head movements for an arbitrary person, we follow the basic idea in \cite{zhou2020makelttalk} and take landmarks as the intermediate representation to present the lip movement and head pose. Following \cite{zhou2020makelttalk}, the lip movements are represented as shifts of key points. Different from \cite{zhou2020makelttalk} that takes the head movement as key points' shifts, we treat the head movements as rotations, which allows the model to predict more natural head poses. Furthermore, as both the speech synthesis and video generation are considered in this work, to simplify the pipeline, the driven speech in the landmark prediction module is presented as Mel-spectrograms that same with the intermediate representation in the TTS system.

\section{Method}
\label{sc:method} 

\subsection{Overall framework}

The overall framework of the proposed method is shown in Fig. \ref{fig:Overall_framework}. The input of this framework is a text and a still face image of a person, and the output is a talking head video of this person where the person speaks the text with a voice that is conditioned on the face image. The proposed framework consists of two sub-modules, i.e., a speech synthesis module and a video generation module. The speech synthesis module is a zero-shot multi-speaker TTS model, with text and a face embedding vector as input. This face embedding vector, which is to provide speaker identity information, is obtained via a pre-trained face encoder (see Section \ref{sc:face_encoder}). 

The video generation module is a speech-driven talking-head video generation module, which is decomposed into two steps: landmark prediction and video generation. In the first step, we generate a sequence of facial landmarks with the synthesized speech intermediate representations, i.e, Mel-spectrograms, and the initial facial landmarks extracted from the input image as input. Then, with the generated landmarks and the input face image, we can generate a sequence of photo-realistic images, and the image sequence is then converted to the final talking head video. Here, an off-the-shelf face 3D landmark detector \cite{bulat2017far} is used to extract the facial landmarks.

\begin{figure}[htbp]
    \centering
    \includegraphics[width=\linewidth]{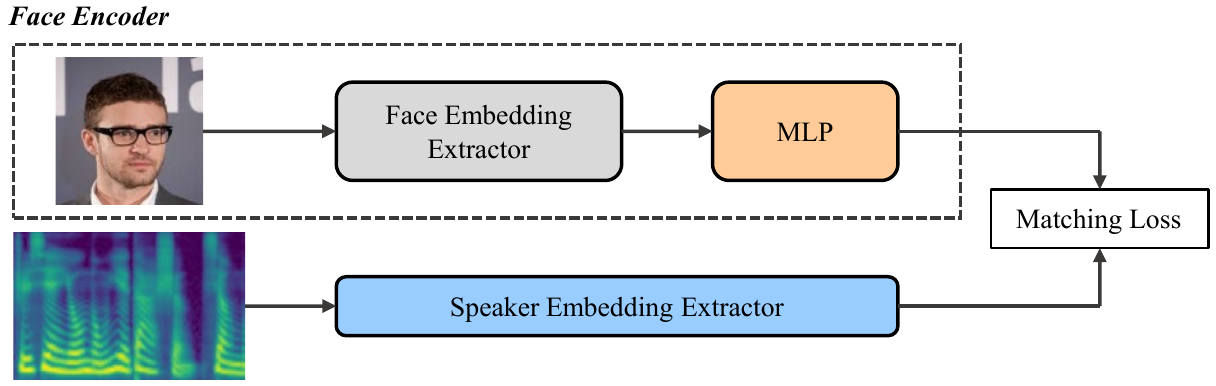}
    \caption{Framework to train the face encoder.}
    \label{fig:face_encoder}
\end{figure}

\subsection{Face Encoder}
\label{sc:face_encoder}

The face encoder is to encode the face image into an embedding vector that provides speaker information in the multi-speaker TTS system (see Section \ref{sc:TTS}). Training such a face encoder could intuitively be done together with the whole multi-speaker TTS system; however, in order to do so a speech database paired with the speaker's face images is needed. Unfortunately, no such database for multi-speaker TTS is available. Fortunately, several speech-visual (talking video with face 
image frames) paired databases exist, which are original collected for, e.g., speaker verification \cite{nagrani2020voxceleb,chung2018voxceleb2} or lip-reading \cite{chung2016lip,chung2017lip}. These databases allow us to train the face encoder separately from the TTS system. Specifically, with a pre-trained speaker embedding extract network that with speech as input, the face encoder can be trained in a teacher-student way with these speech-visual paired databases, i.e., using the speaker embedding extracted from the pre-trained speaker embedding extract network as supervise information to train the face encoder. In this way, ideally, the face embedding and speaker embedding from the same person can represent same information, i.e., speaker identity, so that we can use the face embedding to replace the speaker embedding in a speaker-embedding-based TTS system (see Section \ref{sc:TTS}).

In a typical multi-speaker TTS system \cite{cai2020speaker}, speaker information can be provided by the speaker embedding extracted by a speaker encoder that is trained in a speaker verification task with speech as input. Here, we also trained a speech-based speaker encoder in the speaker verification task with the large margin softmax loss \cite{liu2019large}. Then this speaker encoder works as the teacher to supervise the training of the face encoder. This pre-trained speaker encoder is named as speaker embedding extractor as shown in Fig. \ref{fig:face_encoder} that illustrates the framework to train the face encoder. The model architecture of the speaker embedding network is based on the ResNet-34 \cite{he2016deep} as that in \cite{chung2018voxceleb2}. Here, the last fully connected layer is dropped, and the output speaker embedding is represented as a 1024-D vector with L2 normalization. 

\textbf{Architecture of the Face encoder}. The face encoder consists of an off-the-shelf face embedding extractor that with face image as input\footnote{https://github.com/timesler/facenet-pytorch} \cite{schroff2015facenet} and an MLP block with two linear transformation layers. The output of the face embedding extractor is a 512-D vector. The hidden unit size of the MLP is 2048, and the output size is the same as that of the speaker embedding vector which is 1024. The L2 normalization layer is also added after the MLP in the face encoder as in the speaker embedding extractor. 

\textbf{Training}. 
As the goal is to project a face image into the matched speech embedding space, i.e., to minimize the distance between a matched face embedding and speech embedding pair, the Masked Margin Softmax (MMS) \cite{ilharco2019large} that is designed for visually grounded speech representation learning is adopted as the matching loss. During the training process, parameters of the face embedding extractor and speaker embedding extractor are fixed, and only parameters of the MLP are updated. With the trained image encoder, which consists of the off-the-shelf face embedding extractor and trained MLP layers, we can obtain the final face embedding that is used to replace the speaker embedding in the TTS system.

\begin{figure}[htbp]
    \centering
    \includegraphics[width=\linewidth]{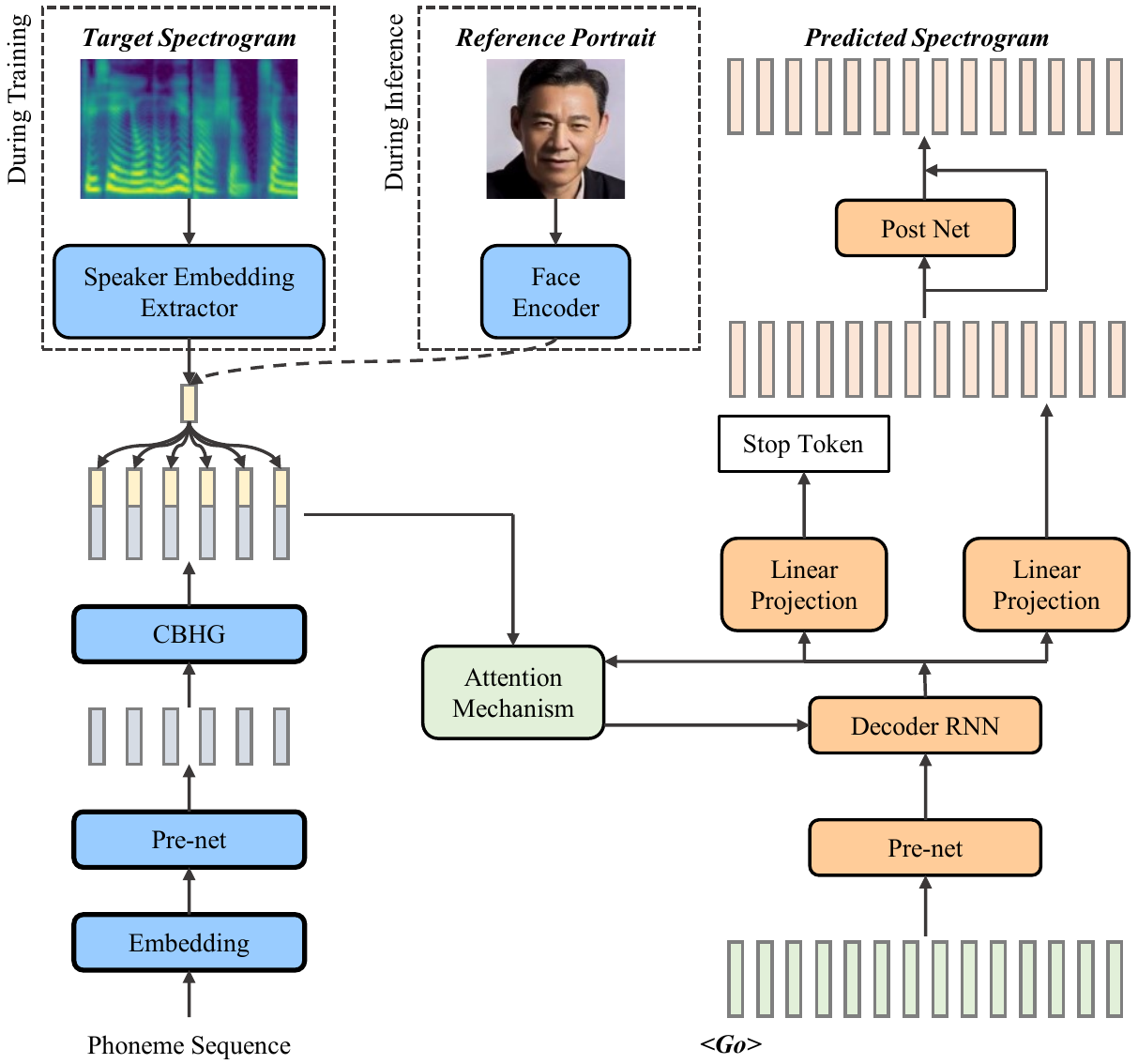}
    \caption{Framework of the face-conditioned multi-speaker TTS.}
    \label{fig:TTS}
\end{figure}

\subsection{Face-conditioned multi-speaker TTS}
\label{sc:TTS}
As mentioned in the last subsection, due to the inaccessible of speech-face paired database, we can not train a face-conditioned multi-speaker TTS directly. Instead, we train a multi-speaker TTS model with text-speech paired database as that in general multi-speaker TTS method \cite{cai2020speaker}. Specifically, during the training process, speaker information can be provided by the speaker embedding extracted with a speaker encoder. Here, the pre-trained speaker embedding extractor shown in Fig.~\ref{fig:face_encoder} is adopted to extract the speaker embedding during the training process. Because the face encoder in Fig.~\ref{fig:face_encoder} is trained with the supervision from speaker embedding, it can be ideally considered that the face embedding and the speaker embedding of the same person share the same representation. Therefore, during the inference processing, we can use the face embedding to replace the speaker embedding, and thus to realized the face-conditioned multi-speaker TTS.

The proposed face-conditioned multi-speaker TTS framework is shown in Fig. \ref{fig:TTS}. The Tacotron-based models \cite{wang2017tacotron,shen2018natural} is used as the Mel-spectrogram prediction model. Specifically, the multi-speaker TTS model has a typical attention mechanism-guided encoder-decoder architecture. The encoder (the left part in Fig.~\ref{fig:TTS}) follows the Tacotron's \cite{wang2017tacotron} encoder that consists of a pre-net and a CBHG block. The text that works as the input to the encoder is represented as a sequence of phonemes that are then embedded into a vector sequence. The decoder (the right part in Fig.~\ref{fig:TTS}) follows the Tacotron2's \cite{shen2018natural} decoder that consists of a pre-net, RNN decoder, and a post-net. Besides, between the RNN decoder and post-net, two linear projection layers are used to predict Mel-spectrograms and stop tokens, respectively. The attention mechanism is to provide a soft alignment between the encoder states and the target Mel-spectrograms. Here, the GMMV2b attention mechanism \cite{battenberg2020location}, which shows better robustness on inferring long utterances than the location-sensitive attention mechanism adopted in Tacotron2 \cite{shen2018natural}, is adopted. The predicted Mel-spectrograms from the decoder are then fed to the Griffin-Lim reconstruction algorithm \cite{griffin1984signal} to synthesize the waveform.

Following the speaker embedding based multi-speaker TTS model \cite{cai2020speaker}, the speaker embedding in the training phase is engaged after the CBHG block. It works as a speaker attribute to provide speaker information for the TTS system. During the inferring phase, the speaker attribute is provided by the face embedding, so that we can synthesize speech guided by the portrait. The standard training method of Tacotron2 \cite{shen2018natural} is adopted to train the face-conditioned multi-speaker TTS.

\subsection{Talking head generation}
The talking head generation part is to generate the talking head video given the Mel-spectrograms synthesized by the TTS module. This process consists of two steps: 1) Mel-spectorgrams-to-facial landmark sequence generation (Section \ref{sc:landmark_prediction}), and 2) landmark sequence-to-video generation (Section \ref{sc:photo_generation}). The landmark sequence generation module is also designed as an encoder-decoder architecture. As shown in Fig.~\ref{fig:Overall_framework}, there are three encoders in this landmark generation (prediction) module, i.e., audio encoder, landmark encoder, and quaterion encoder, which are for the encoding of synthesized Mel-spectrograms, facial landmark of input image, and orientation of the face in the input image, respectively. After concatenating, the output from these three encoders are input to the decoder to generate the facial landmark sequence. By connecting consecutive key points of facial landmarks in each frame with pre-defined colors, i.e., using different colors to distinguish different parts as that in \cite{zhou2020makelttalk}, we can get a sequence of facial sketches. These facial sketches are then concatenated with the input face image, resulting in a sequence of 6-channel images used to generate photo-realistic frames in the final video with an image-to-image translation way. 

\subsubsection{Landmark generation}
\label{sc:landmark_prediction}
The facial landmark generation module follows the basic idea of MakeItTalk~\cite{zhou2020makelttalk}, i.e., separately predicting the lip movement and head movement, and combing them to generation the final facial landmarks. Compared to \cite{zhou2020makelttalk}, there are mainly two differences in our work: 1) Instead of treating the head movement as facial landmarks' shift, we treat the head movement as head rotation. This rotation parameter is represented by quaternions. Therefore, three encoders, including audio encoder, landmark encoder, and quaternion encoder, are included in this module; 2) Instead of using bi-directional LSTM as in \cite{zhou2020makelttalk} or time-delay LSTM as in other related work \cite{yu2020multimodal,suwajanakorn2017synthesizing} to predict landmarks, a CNN-based block is adopted before the LSTM layer to make a frame get more contextual information from adjacent frames.
  
A vivid talking head should not only have synchronized lip movement but also natural head pose movements. While the lip movements and facial expressions, e.g., the jaw and eye movements, are performed in a 3D space, the final landmarks are drawn on a 2D plane, i.e., facial sketch, to render the photo-realistic facial image. Therefore, movements in the direction that is perpendicular to the face are not important for the landmark-to-image generation. In contrast, rotations of the head (referred to as head pose) are performed in 3D, and even drawn on a 2D plane, different head poses would lead to different head sketch on the plane. To effectively model facial expressions and head poses, we decompose the landmark prediction into landmark shift in a 2D plane and head rotations in a 3D space with the help of 3D facial landmark detector \cite{bulat2017far} that can detect 3D coordinates of landmarks from images (video frames). 

Given an input face image $I$, the extracted facial landmarks consist of 68 key points, each of which is represented by three-dimensional coordinate values. To capture the facial expression-related movements, such as the lip and jaw movements, in the same plane, we first register the facial landmarks to a front-facing standard facial template as done in \cite{zhou2020makelttalk} with the ICP method proposed in \cite{segal2009generalized}. The orientation of the original face is represented as a set of quaternion numbers $q \in {R^4}$. The landmarks of the input image is important conditional information for the prediction of landmarks. As shown in Fig. \ref{fig:Overall_framework}, Mel-spectorgrams, the frontal facial landmarks, and quaternions are encoded by the audio encoder, landmark encoder, and quaternion encoder, respectively. Outputs from these three encoders are concatenated to work as input to the landmark decoder that generates the new landmarks. 

We denote the sequence of mel-spectrograms as $S = \left\{ {{s_1},{s_2},...,{s_T}} \right\}$, where $T$ is the sequence length. The goal is to generate a sequence of frontal landmarks $\hat P = \left\{ {{{\hat p}_1},{{\hat p}_2},...,{{\hat p}_T}} \right\}$ and a sequence of quaternions $\hat Q = \left\{ {{{\hat q}_1},{{\hat q}_2},...,{{\hat q}_T}} \right\}$. The corresponding ground-truth landmark sequence and quaternion sequence are $P = \left\{ {{p_1},{p_2},...,{p_T}} \right\}$ and $Q = \left\{ {{q_1},{q_2},...,{q_T}} \right\}$, respectively. In practice, we drop the dimension of the landmarks in the depth direction (z-axis value), and only 2D landmarks are used to presented landmark displacements, which means each point is represented by two-dimensional coordinates, i.e., \textit{<x-axis value, y-axis value>}. Considering 68 key points of facial landmarks, one landmark frame can be represented as $p_t \in {R^{136}}$ by concatenating the coordinate values of x-axis and y-axis.

Due to that the face shape various with different persons, making it challenging to predict landmarks for a new person that was never seen during the training process. To face this challenging, in \cite{zhou2020makelttalk}, instead of directly predicting the landmarks, they predict the displacements of landmarks, and these displacements are added to the base landmarks' coordinates. In this paper, we also take this strategy to the arbitrary talking head generation. To this end, we have to to choose a frame to provide the base landmarks and quaternions, which are referred to as base landmarks and base quaternions hereafter. During training, the training samples pair is a sequence of Mel-spectorgrams and a sequence of landmarks extracted from the video, so that we can randomly choose one landmark frame to provide the base facial landmarks and quaterions. During the inference process, only one input image is available, and the base facial landmarks and quaterions are provided by this input image. The prediction of the landmarks' displacements can be formulated as:
\begin{equation}
\begin{array}{l}
\begin{aligned}
& {s'_i} = AE\left( {{s_i};{{\text{w}}_{AE}}} \right) \\
& p' = LE\left( {p_{init};{{\text{w}}_{LE}}} \right) \\
& q' = QE\left( {q_{init};{{\text{w}}_{QE}}} \right) \\
& {m_i} = {\text{concat}}\left({{s'_i}, p',q'} \right),M = \left\{ {{m_1},{m_2},...,{m_t}} \right\} \\
& \Delta P, \Delta Q' = LD\left( {M;{{\text{w}}_{LD}}} \right) \\
&  Q = LSTM\left( {\Delta Q';{{\text{w}}_{LSTM}}} \right),
 \end{aligned}
\end{array}
\end{equation}
where $AE$, $LE$, $QE$, and $LD$ are the audio encoder, landmark encoder, quaternion encoder, and landmark decoder, respectively. ${\text{w}}_{AE}$, ${\text{w}}_{LE}$, ${\text{w}}_{QE}$, ${\text{w}}_{LD}$, and ${\text{w}}_{LSTM}$ are learnable parameters. $M$ is a sequence of features, each of which is obtained by concatenating the frame-level speech embedding vector ${s'_i}$, base landmark embedding vector $p' $, and base quaternion embedding vector $q'$. The output of the landmark decoder is a sequence of concatenated landmark displacements and preliminary quaternion changes in the frame level. Specifically, each frame of the decoder's output is a 140 dimensional vector that consists of 136 dimensions for landmark displacements and other 4 dimensions for quaternion changes. After amputating each frame, we can get a sequence of landmark displacements $\Delta P$, and a sequence of quaternion changes $\Delta Q'$. Compared to lip movements, the head pose changes more slowly and smoothly. To make the predicted head pose changes be smooth, a further $LSTM$ is adopted to deal with the predicted quaternion changes $\Delta Q'$, and result in the final quaternion changes $\Delta Q$ . Formulaically, the predicted frame-level frontal landmarks and quaternions can be obtained via:
\begin{equation}
\begin{array}{l}
\begin{aligned}
& {\bar p_t} = p_{init} + \Delta {p_i} \\
& {\bar q_t} = q_{init} + \Delta {q_i}. 
\end{aligned}
\end{array}
\end{equation}
With the predicted quaternions, we can get the rotation matrix $M$, with which we can get the final rotated landmarks:
\begin{equation}
    \left[ {{\hat p_i},e} \right] = \left[ {\bar p_i,e} \right] \cdot M
\end{equation}
where $e$ is a unit vector.

\textbf{Model architecture}.
All the designed encoders, i.e., audio encoder, landmark encoder, and quaternion encoder, are multi-layer perceptrons (MLP) with two linear transformation layers, where the first linear transformation layer is followed by a layer normalization \cite{ba2016layer} and an activation function of LeakyReLu \cite{maas2013rectifier}. The hidden unit sizes of $AE$, $LE$, and $QE$ are 512, 256, and 64, respectively. The vector dimensions of ${s'_i}$, $p'$, and $q'$ are 512, 128, and 4 respectively. Therefore, the dimension of $m_i$ is 644.

The landmark decoder $LD$ consists of a 1D-CNN block, a bidirectional LSTM block, and an MLP. The 1D-CNN consists of six 1-D convolutional layers with unit sizes of 512, 512, 1024, 1024, 1024, and 2048, respectively. Instance normalization is used after the first convolutional layer, while the other convolutional layers are followed by batch normalization. The MLP follows the same structure as those encoders, with a hidden unit size of 512.

\textbf{Objective function}. The objective functions for the displacement prediction consist of an $L_2$ regression loss and a pairwise inter-frame loss. Specifically, the $L_2$ regression loss is defined as:
\begin{equation}
    { {\cal L}_d} = \sum\limits_{t = 1}^T {\sum\limits_{i = 1}^N {\left\| {{p_{i,t}} - {{\hat p}_{i,t}}} \right\|_2^2} } 
\end{equation}
where $N$ is the batch size. The pairwise inter-frame loss is defined as:
\begin{equation}
    { {\cal L}_{in}} = \sum\limits_{t = 2}^T {\sum\limits_{i = 1}^N {\left\| {\left( {{p_{i,t}} - {p_{i,t - 1}}} \right) - \left( {{{\hat p}_{i,t}} - {{\hat p}_{i,t - 1}}} \right)} \right\|_2^2} }.
\end{equation}

The objective function for the quaternion prediction is a $L_1$ regression loss:
\begin{equation}
    {{\cal L}_q} = \sum\limits_{t = 2}^T {\sum\limits_{i = 1}^N {\left\| {{q_{i,t}} - {{\hat q}_{i,t}}} \right\|} }.
\end{equation}

The total loss function of the landmark prediction is
\begin{equation}
    {{\cal L}_L} = {{\cal L}_d} + {{\cal L}_{din}} + {{\cal L}_q}.
\end{equation}

\subsection{Landmark to photo-realistic image}
\label{sc:photo_generation}

In the generated landmark sequence, each frame consists of facial landmarks with a special head pose and lip shape. With facial landmarks of each frame, we can generate the photo-realistic face image by the face generator in Fig. \ref{fig:Overall_framework}. Here we take the UNet architecture from \cite{esser2018variational,zakharov2019few,zhou2020makelttalk} as the face generator to perform this landmark-to-image translation. The landmarks of each frame are drawn as a portrait sketch on a 2D plane by connecting the key points with pre-defined colorful lines, as shown in Fig. \ref{fig:Overall_framework}. Then this portrait sketch is concatenated with the input image, resulting in a 6-channel image with a resolution of $256 \times 256$ which will work as the input to the face generator. The output is a photo-realistic face image that with the same facial key points as input landmarks.

To train the image generator, in addition to minimizing the L1 pixel-level distance and perceptual feature distance between the reconstructed face and the training target face as in \cite{zhou2020makelttalk}, conditional generative adversarial training loss in \cite{isola2017image} is also used. Following \cite{isola2017image}, the discriminator is a patch-based fully convolutional network. The input of the discriminator is also the channel-wise concatenation of the portrait sketch and the input image (real) or the generated image (fake).

\section{Experiments and Results}
\label{sc:experiments}

\subsection{Database}

\begin{table*}[]
\centering
\setlength{\tabcolsep}{4mm}
\caption{Databases that were used to train the different modules.}
\begin{tabular}{lcccl}
\toprule
Database             & Adopted Modality & Language & Speaker number & \multicolumn{1}{c}{Used for which module} \\ \midrule
AISHELL-3            & Text-Audio       & Mandarin & 218            & TTS                                       \\
VCTK                 & Text-Audio       & English  & 110            & TTS                                       \\
Aidatatang-200zh     & Audio            & Mandarin & 600            & Speaker embdding extractor                \\
VoxCeleb2 subset \cite{chung2018voxceleb2}    & Audio-Video      & English  & 433            & Face encoder; Image translation model     \\
Cn-Celeb subset \cite{li2020cn}     & Audio-Image      & Mandarin & 313            & Face encoder                              \\
Obama Weekly Address \cite{suwajanakorn2017synthesizing} & Audio-Video      & English  & 1              & Landmark prediction model                 \\ \bottomrule
\label{tb:dataset}
\end{tabular}
\end{table*}

Table \ref{tb:dataset} lists the various databases that were used to train the different modules of the proposed method. In addition to these databases, we also collected several data to evaluate the face-conditioned multi-speaker TTS and the final generated talking head video. These collected data will be introduced in Section \ref{sc:evaluation}. These databases will be introduced below grouped by the corresponding module.

\subsubsection{Database for the TTS} In order to be able to make both Mandarin and English speaking talking head videos,  databases of both languages, i.e., AISHELL-3\footnote{http://www.aishelltech.com/aishell\_3
} and VCTK\footnote{https://datashare.ed.ac.uk/handle/10283/3443} were adopted to train the multi-speaker TTS model. AISHELL-3 is a multi-speaker Mandarin speech database with speech by 218 native Chinese Mandarin speakers with a total of 88,035 utterances. VCTK is a multi-speaker English speech database with speech from 110 English speakers with various accents, where each speaker reads out around 400 sentences. The multi-speaker model is trained with the these two databases together, which allows the trained model to be used for both Chinese and English. Note that these two databases are only used for the training of the multi-speaker TTS model. In the final speech synthesis for the talking head video, the speaker identify is provided by a face image. However, no paired face image exists in AISHELL-3 and VCTK. Therefore, only 100 transcriptions are randomly selected as the test sentences for the whole talking head generation task, and their paired utterances are not used.

\subsubsection{Database for face encoder}
\label{sc:database for face encoder}
The speaker embedding extractor which works as the teacher to train the face encoder is trained with the database of Aidatatang-200zh\footnote{http://www.openslr.org/62/}. This is a Chinese Mandarin speech corpus that contains 200 hours of speech data from 600 speakers. After obtaining the pre-trained speech embedding extractor, databases which pair speech and faces are needed to train the face encoder. We use two subsets from VoxCeleb2\footnote{http://www.openslr.org/49/} and Cn-Celeb\footnote{http://www.openslr.org/82/} to train the face encoder. 

Both VoxCeleb2 and CN-Celeb were originally designed for the task of speaker verification. VoxCeleb2 is an audio-visual database, which consists of short clips of human speech extracted from interview videos uploaded to YouTube. The associated video track provides us the matched face images to the corresponding utterances. Here, a subset of VoxCeleb2 \cite{siarohin2019first} is adopted. This subset consists of 16128 English utterances uttered by 433 speakers. Following \cite{siarohin2019first}, 422 speakers with 15729 utterances are used as training data and other 11 speakers are used as the test set to provide speaker image in the talking head generation task. For each speaker, we randomly extracted 50 frames from their talking videos to build a paired face database. 

The original Cn-Celeb contains more than 130,000 utterances from 1,000 Chinese celebrities, but without face information. To obtain the speech-image pairs, we collected a face image database of a part of the speaker identities in the Cn-Celeb. Specifically, this collected face database consists of 313 speakers and each speaker has 40 to 100 face images downloaded from Baidu Image\footnote{https://image.baidu.com/}. Therefore, the final database to train the face encoder consists of 735 speakers and 28450 utterances. 

\subsubsection{Database for the talking head generation}
Following \cite{zhou2020makelttalk}, the Obama Weekly Address database \cite{sadoughi2017synthesizing}, which contains around 6 hours of Obama's speeches, is used to train the landmark prediction model. We cut the audio signals into fixed-length utterances with the duration of 3s. Subsequently, the utterances are split as 90\%, 5\%, and 5\% for training, validation and test, respectively. 

The database to train the image translation model is the subset of VoxCeleb2 introduced in Section \ref{sc:database for face encoder}. Different from the data pairs used in Section \ref{sc:database for face encoder}, which are speech-image pairs, here, speech-video pairs are used.

\subsubsection{Data processing}

In all proposed modules, speech is represented as Mel-spectrograms with the same parameters. Specifically, the Mel-spectrograms are computed through a short-time Fourier transform (STFT) with 50 ms frame size and 12.5 ms frame hop, resulting in a frame frequency of 80Hz. The frame rate of the videos from the Obama Weekly Address database is 25 fps. To align the Mel-spectrograms and video frames, we up-sample the video frame rate to 80 fps. This up-sampling is performed on the landmark features instead of on the raw video frames.

\subsection{Evaluation}
\label{sc:evaluation}
The goal of our task is to generate voiced talking head video with text and the face image as input. A good generated result should consists of: 1) reasonable speech that is likely produced by the person in the given face image, and 2) the video is synchronized with synthetic speech. Therefore we have to evaluate synthetic speech and the generated video respectively. 


\subsubsection{Face-conditioned multi-speaker TTS}
\label{sc:user_study_TTS}
The goal of the face-conditioned multi-speaker TTS is to synthesize speech that sounds like it could be produced by the given face image. This makes the evaluation a subjective task. Therefore, a human perceptual rating experiment is performed to evaluate the synthesized speech. However, when there is no reference speech, it is very hard for participants to rate synthetic speech only based on the given face image. To make this evaluation easier for participants, instead of rating a score for given speech, participants are asked to choose the better one from two compared samples, which is called as A/B test. Here, two of A/B tests are performed with different compared speech. In one, assuming that a speaker's synthetic speech, which is synthesized based on real speech of this person to provide speaker information, can be treated as ground-truth synthetic speech, our goal is to test whether our face image-based synthetic speech can achieve comparable results compared to this reference speech-based synthetic speech. Therefore, in this A/B test, a pair of compared synthetic samples are from the reference speech-based multi-speaker TTS method and the face image-based multi-speaker TTS method respectively. In the other, our goal is to test whether our face image-based results are obviously superior to synthesized speech that is conditioned on reference speech that is randomly selected from the training set with the same gender. Hence, in this second test, the compared samples are synthesized by the reference speech-based multi-speaker TTS, in which reference speech is randomly selected with the prior Knowledge of gender.

In each A/B test, we give 16 groups of samples. Each group consists of a face image, a speech sample synthesized by our face image-based multi-speaker TTS, and a compared speech sample synthesized by the reference speech-based multi-speaker TTS (the reference is ground-truth speech of the person in this given image or randomly selected from training set but with the same gender). Both synthetic speech samples are with the same textual sentences, and the participants are asked to choose the one that they think is more likely produced by the identity in the given face image. A third choice that ``They are similar'' is also an option, which allows participants to make their decision when they can not tell which one is better.

Both VoxCeleb2 and Cn-Celeb databases that used to train the face encoder are collected from persons who are celebrities. These celebrities may familiar to the participants, which could influence the judgment due to the prior knowledge of celebrities' voices. Therefore, we do not use the identities from these two databases as evaluation data. Instead, we collected 16 (8 men and 8 women) talking videos recorded by unknown YouTubers from YouTube\footnote{https://www.youtube.com/}. Because all the participants are Chinese native speakers, the collected 16 videos contain recordings by Chinese speakers. Speech from these collected videos allow us to synthesize speech with the reference speech-based multi-speaker TTS method, which works as the compared method in one of the two A/B test experiments. The text used as input to our model are taken from the test set of AISHELL-3. In the human perceptual rating experiments, a total of 27 people (8 females and 19 males with age range of 18 to 40) participated.

\subsubsection{Talking head generation}

We decompose the speech-to-video generation into speech-to-landmark generation and landmark-to-video two stages. For the speech-to-landmark generation, as the ground-truth landmark sequences are available from the test set of Obama Weekly Address database, objective evaluations are performed to compare the generation landmark sequences and ground-truth landmark sequences. For the final generated videos, human perceptual rating experiments are performed to evaluation the naturalness of the video and also the synchronization of the lip movements and speech.

\textbf{Evaluation Metrics}. Following \cite{zhou2020makelttalk}, we evaluate the talking head generation and particularly the accuracy of the lip movements using the landmark distance for lips (\textbf{D-LL}), landmark velocity difference for lips (\textbf{D-VL}), and difference in the open mouth area (\textbf{D-A}) as evaluation metrics. D-LL is the average Euclidean distance between the predicted lip landmarks and the ground-truth ones. D-VL is the average Euclidean distance between the predicted lip landmark velocities and that of the ground-truth ones. D-A represents the average difference between the area of the predicted mouth shape and the ground-truth one.

\textbf{User study}. Given a generated video, participants are asked to rate the video in terms of 1) the synchronization of the lip movements and speech, and 2) the overall realness of the video, respectively, on a 5-point scale using the slider. A score of 1 means "very bad" and a score of "5" means excellent. In this experiment, twenty face images are randomly collected from Google Image to generate the talking head videos. Ten of them are for Chinese talking head videos and others are for English talking head videos. For each language, two sentences are randomly selected from AISHELL-3 or VCTK to work as the input sentences. Note that, the cross-lingual task is not considered in this paper. During the inference process, the language is manually defined based on whether the person in the image is Chinese or not. In this user study, a total of 22 people (5 females and 17 males with age range of 18 to 40) participated.

\subsection{Results}
In this section, the evaluation results for 1) face-conditioned multi-speaker TTS to test whether this method can produce synthetic speech that is likely produced by the person from the given image; 2) talking head generation to test the synchronization and naturalness of the generated video, are presented. In addition, we also visualize the generated video frames with non-real person face image, e.g., cartoons and statues, as input in this section.

\subsubsection{Face-conditioned multi-speaker TTS.} 
\begin{figure}[tbp]
    \centering
    \includegraphics[width=\linewidth]{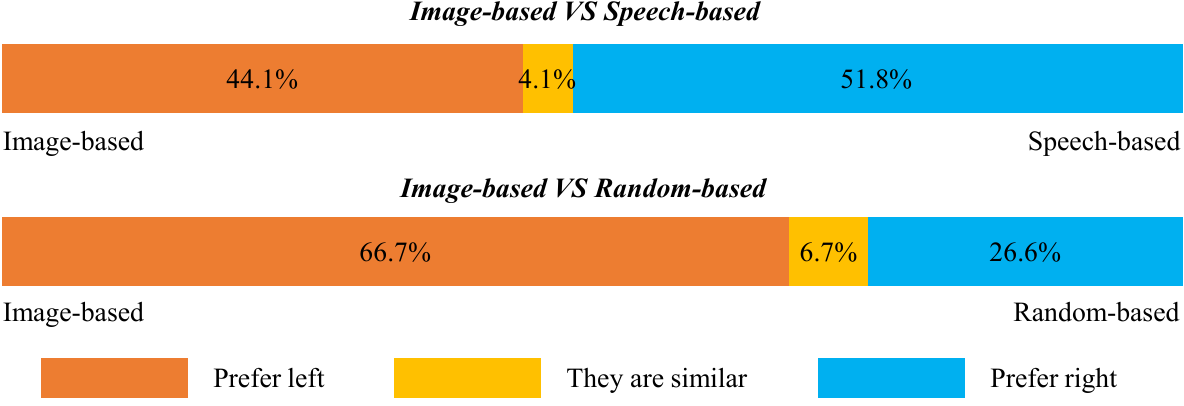}
    \caption{The results of the user study for the evaluation of the face-conditioned multi-speaker TTS.}
    \label{fig:user_study_TTS}
\end{figure}

\begin{figure}[tbp]
    \centering
    \includegraphics[width=\linewidth]{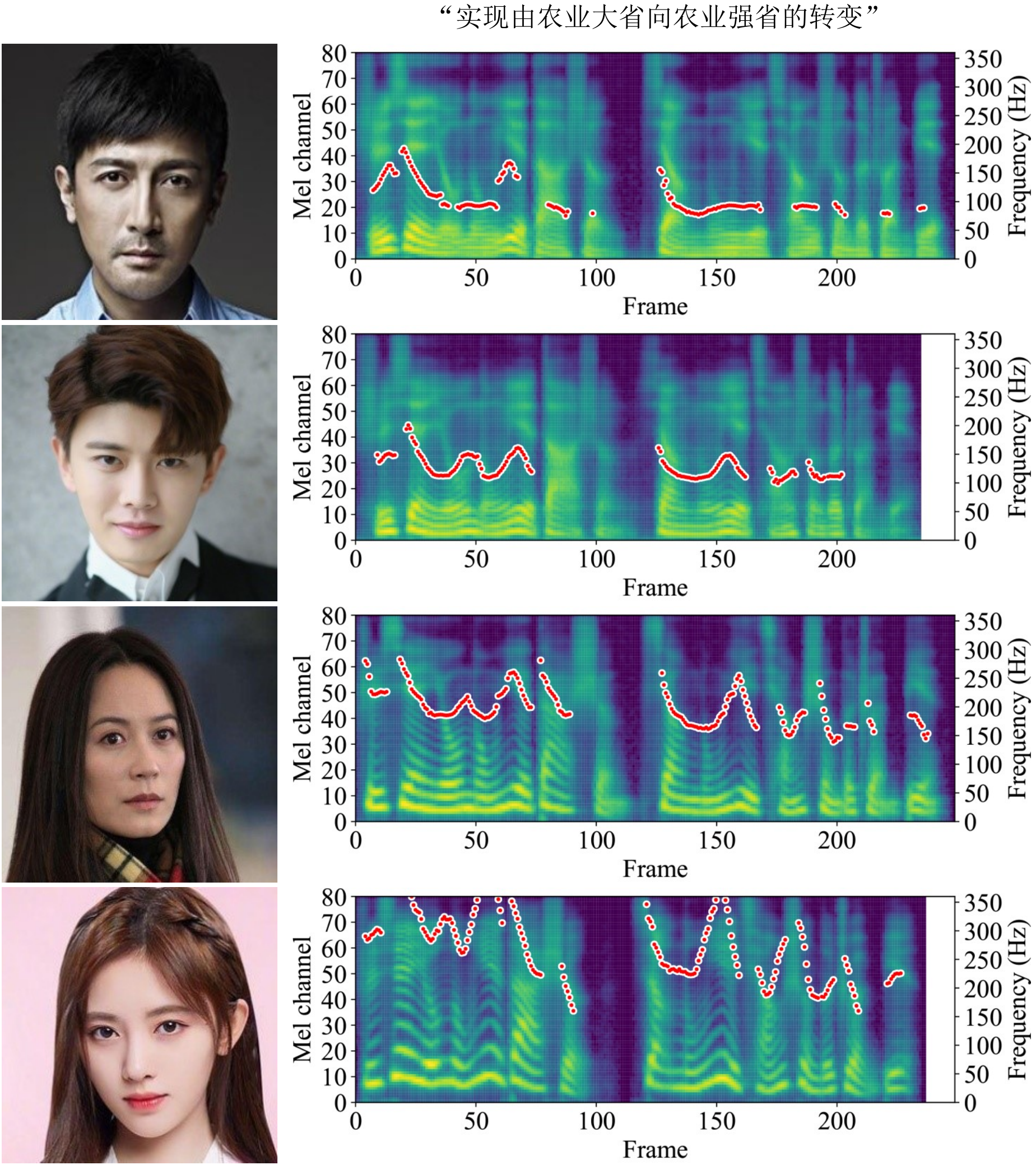}
    \caption{Examples of the synthesized speech (right panels) after conditioning on the face image (left panels).}
    \label{fig:pitch}
\end{figure}

The human perceptual experiment (see Section \ref{sc:user_study_TTS}) results are shown in Fig. \ref{fig:user_study_TTS}, which displays the percentage of the total votes in the two A/B test experiments, respectively. The upper bar shows the results of the reference speech-based method and the face image-based method, in which the reference speech-based method can treated as an upper boundary in the multi-speaker TTS task because the speech embedding vectors are extracted from the ground-truth speech signals. While 7.7\% gap exists between the reference speech-based method and the face image-based method, a one-way ANOVA shows that no significant difference exists between the image-based results and speech-based results (F = 0.17, p = 0.68). Considering that 4.1\% votes are for ``they are similar", nearly half votes are for that the face image-based results are not worse than the reference speech-based results, indicating that our face image-based method can produce reasonable speech according to one's portrait. 

Even though in the A/B test the participants only need to choose one from two given samples, it is still not easy to tell which one is really better, because no explicit relation exists between the portraits and voices, which may make the perceptual comparisons between the reference speech-based results and face image-based results unconvincing. Therefore, we have to know whether the face image-based method is really better than the randomly selected speech-based method. As shown in the other A/B test result, the proposed image-based method was more often chosen as the "better" voice and thus received a much higher vote percentage. A one-way ANOVA also confirmed these results: there is a significant difference in vote number between the image-based results and randomly selected results (F  =   7.25, p = .01). The similar performance with reference speech-based method and significant superiority to the randomly selected speech-based method demonstrates that the proposed face-conditioned multi-speaker TTS model can synthesize reasonable voices according to the input face images. Thus this method can be used to synthesize synchronized speech and talking heads bypassing the dependency on the reference speech.

To present the synthesized results of our face-image based method intuitively, some cases are shown in Fig. \ref{fig:pitch}. In this figure, the input images and their corresponding synthesized spectrograms are presented. Besides, the fundamental frequency of synthesized speech is also drawn on the spectrograms. As can be seen, with the same sentence but different face images as input, the generated spectrograms and also fundamental frequencies are significantly different, indicating that the face image indeed can provide discriminative identity information.  

\begin{figure*}[htbp]
    \centering
    \includegraphics[width=\linewidth]{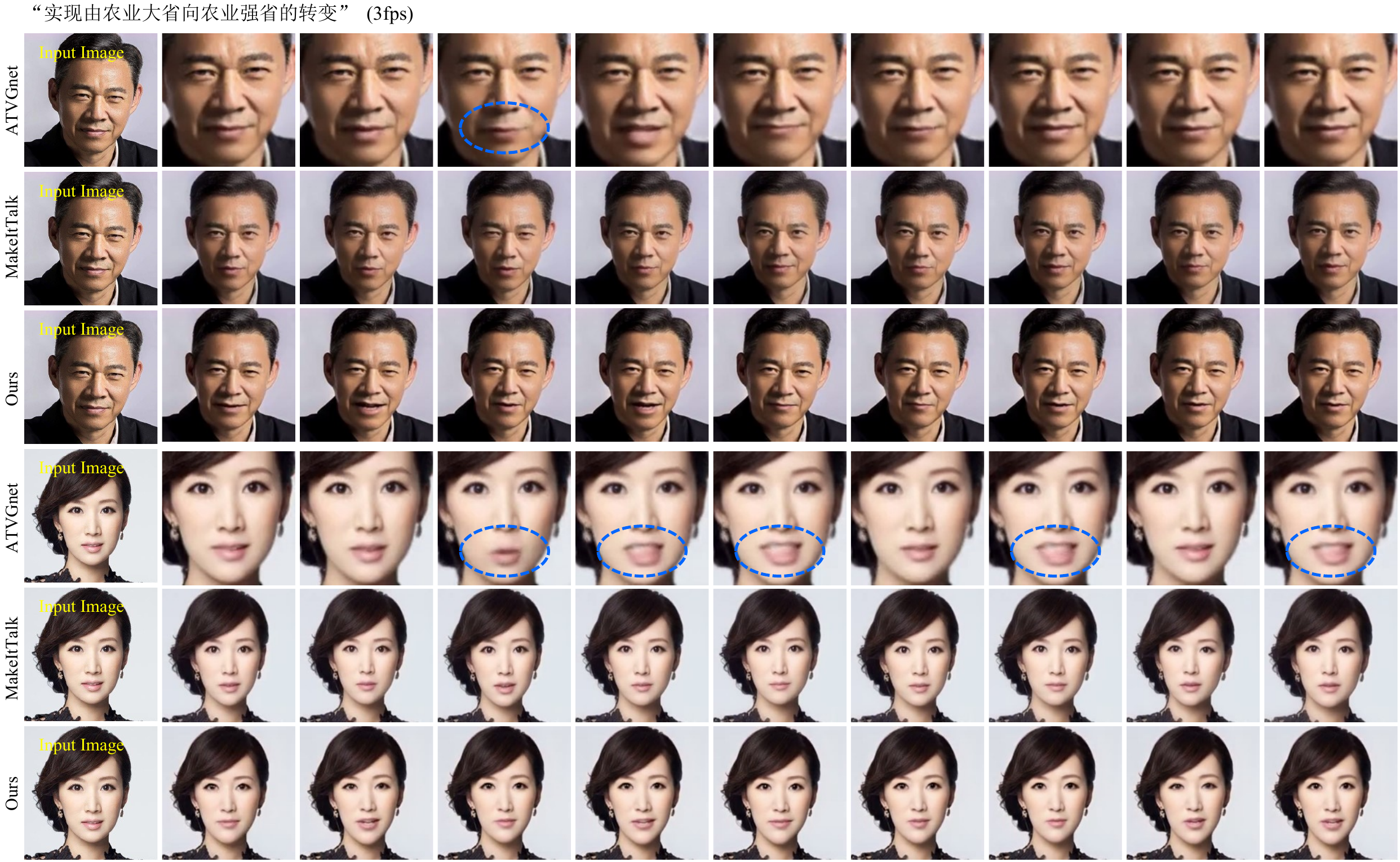}
    \caption{Comparison of talking head generation with target independent audio-driven methods. We recommend readers watch the results in the video demo.}
    \label{fig:visual_comapre}
\end{figure*}

\subsubsection{Talking head generation}
\textbf{Lip movement prediction}. We first evaluate how well the predicted lip landmarks synchronized with the ground-truth lip landmarks and compare the performance of our lip movement prediction to that of MakeItTalk \cite{zhou2020makelttalk}, which is a state-of-the-art landmark-based talking head method for arbitrary persons. MakeItTalk is based on the same 3D landmark extractor as our model is, and its landmark prediction model is also trained on the Obama weekly talking database, just as our model. This allows for a fair comparison between our method and that of MakeItTalk. Moreover, to evaluate the proposed CNN-LSTM-based landmark decoder,  our approach is compared to two variants of the proposed methods: the TDLSTM approach and the BLSTM approach, in which the landmark decoder in \ref{fig:Overall_framework} is replaced by a time-delay LSTM and bi-directional LSTM, respectively, both of which are popular architectures in related work~\cite{zhou2020makelttalk,yu2020multimodal,suwajanakorn2017synthesizing}. The results are shown in Table \ref{tb:lip_eval}. Bold indicates the best result. As can be seen, our method outperforms MakeItTalk in terms of all evaluation metrics. The proposed method also outperforms the TDLSTM-based and BLSTM-based methods, indicating the superiority of the proposed CNN-BLSTM landmark decoder over the TDLSTM and the BLSTM decoders.

\textbf{Video generation}. In terms of the final generated photo-realistic videos, frames of several generated videos are presented in Fig. \ref{fig:visual_comapre}, in which another talking head generation method ATVGnet \cite{chen2019hierarchical} that is designed for arbitrary persons is also compared. Compared to MakeItTalk and our proposed method, ATVGnet crops the face region of the input image and no head pose is considered. While the amplitude of the lip movements is larger for the ATVGnet generated talking faces than that for thoese generated by MakeItTalk and our method, many of them are unnatural, such as those lip regions circled by blue circles. Compared to the input image (left-most column), obvious distortions appear in the results generated by MakeItTalk. For instance, in the first case, the generated results of MakeItTalk (the second row in Fig. \ref{fig:visual_comapre}) show a thinner facial shape than the original facial shape. Besides, there is a loss of the facial details in these generated frames, which reduces the sharpness of the generated faces. In contrast, the facial details are preserved well in the frames generated by our method, and no distortion appears in our generated frames. Quantitative subjective comparisons from the user study experiment are shown in Table \ref{tb:user_study_mos}. Bold indicates the best results. As shown in this Table, our method outperforms ATVGnet and MakeItTalk in items of lip sync quality and the overall realness. A one-way ANOVA with the method as the factor (three levels of the factor are included, i.e., ATVGnet, MakeitTalk, and our method) and rating score as variable shows that significant difference exists for rating different methods, indicating the good performance of the proposed landmark-based audio-driven talking head generation module.

\begin{table}[]
\centering
\setlength{\tabcolsep}{5.5mm}
\caption{Quantitative evaluation of lip landmark predictions. For all evaluation metrics, a lower value means better performance. Bold indicates the best result of each metric.}
\label{tb:lip_eval}
\begin{tabular}{l|ccc}
\toprule
Method     & D-LL$\downarrow$          & D-VL$\downarrow$           & D-A$\downarrow$            \\ \midrule
MakeItTalk & 0.143          & 0.036          & 0.143          \\
TDLSTM     & 0.105          & 0.027          & 0.130          \\
BLSTM      & 0.101          & 0.027          & 0.115          \\
Ours       & \textbf{0.095} & \textbf{0.026} & \textbf{0.105} \\ \bottomrule
\end{tabular}
\end{table}

\begin{table}[]
\centering
\setlength{\tabcolsep}{4mm}
\caption{Mean Opinion Scores (MOS) for the video evolution. Larger is better, and the maximum value is 5. Bold indicates the best result. All p < .001 in a one-way ANOVA.}
\label{tb:user_study_mos}
\begin{tabular}{l|cc}
\toprule
Method \ MOS & Lip Sync Quality & Overall Realness \\ \midrule
ATVGnet      & 2.92             & 2.60             \\
MakeItTalk   & 2.55             & 2.68             \\
Ours         & \textbf{3.16}             & \textbf{3.17}             \\ \bottomrule
\end{tabular}
\end{table}

\begin{figure}[htbp]
    \centering
    \includegraphics[width=\linewidth]{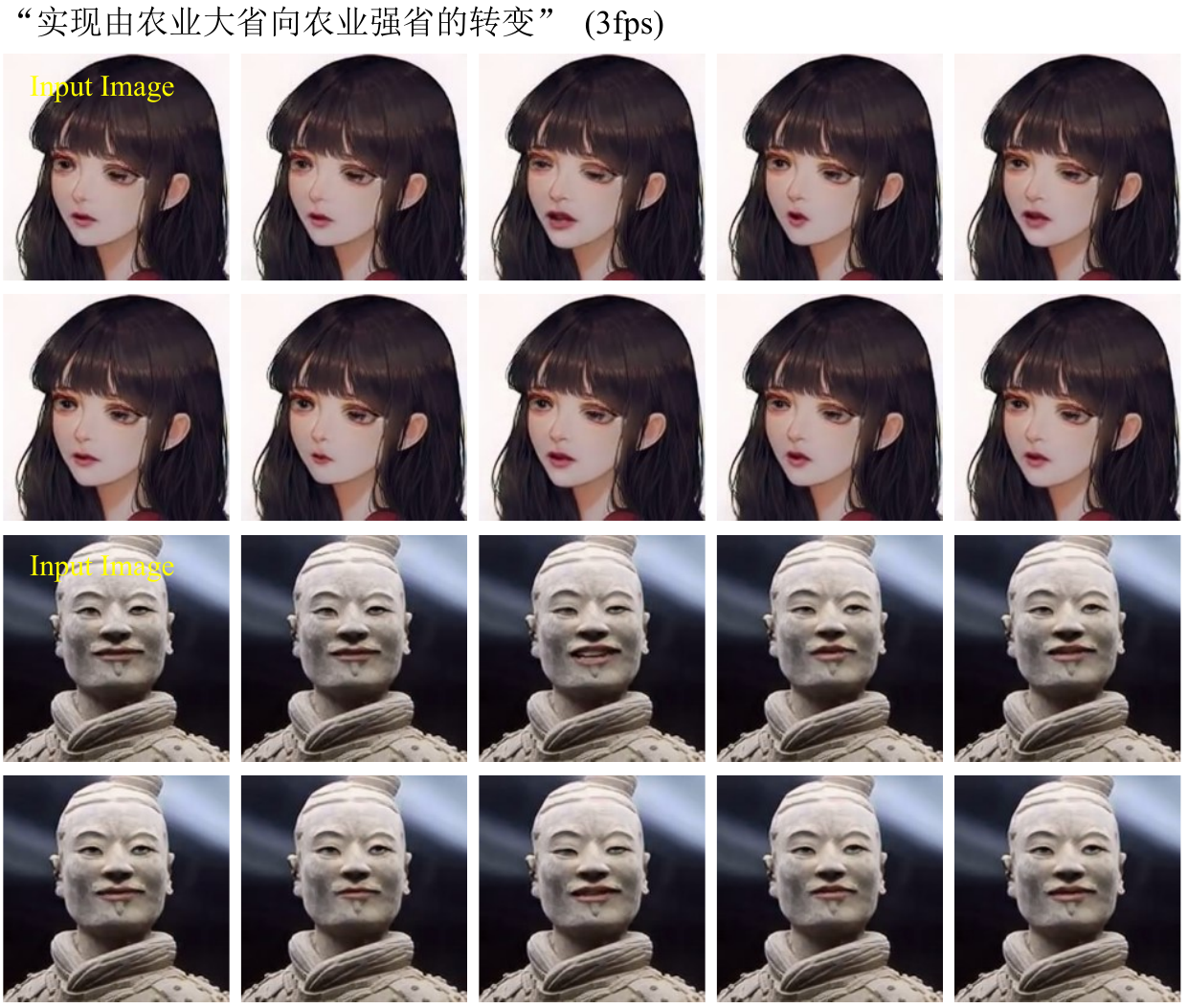}
    \caption{Talking head generation for non-real person portrait.}  \label{fig:non-real video}
\end{figure}

\begin{figure}[htbp]
    \centering
    \includegraphics[width=0.8\linewidth]{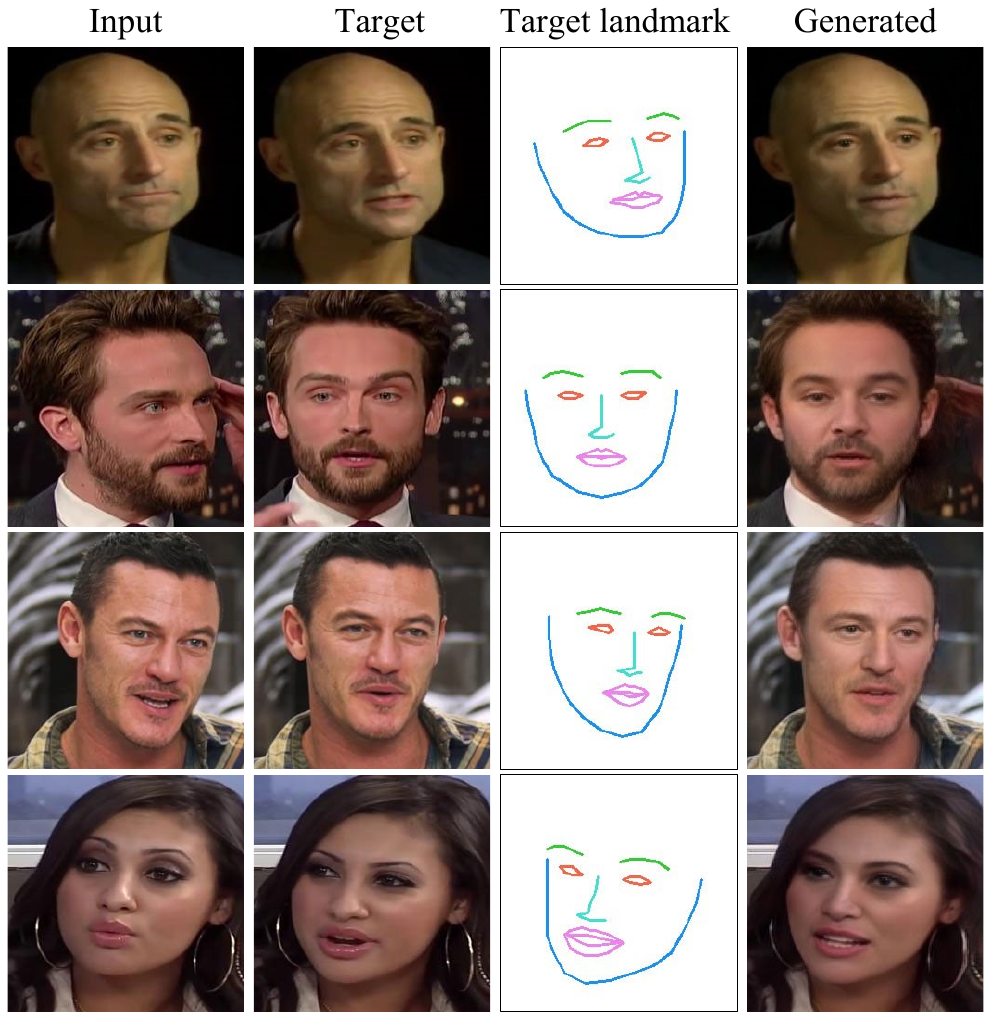}
    \caption{Generated images based on ground-truth landmarks.}
    \label{fig:weak_landmark}
\end{figure}

\textbf{Non-real person talking head generation.} In addition to generating talking head videos with photos of real persons as input, we also present the performance of the proposed method for generating talking faces of non-persons, e.g., cartoons and statues. Frames of two generated videos with a cartoon image  and a state photo as input are shown in Fig. \ref{fig:non-real video}. As can be seen (we also recommend readers watch the demo video), the proposed method is able to generate talking head video for these non-real person face image.

\section{Discussion}
\label{sc:discussion}
In this paper, a voiced talking head generation method is proposed. Different from the previous work that either voiced talking head generation can only be used for one person, or the methods that designed for arbitrary persons do not produce speech, our method, for the first time, allows to generating talking head video for arbitrary persons and meanwhile producing speech that is likely produced by corresponding persons only with the text and the face image as input.

In MakeItTalk \cite{zhou2020makelttalk}, the talking head movements are treated as the shifts of landmarks as that of the lip movements. In contrast, we argue the head movements are more like rotations instead of shifts. In the proposed method, quaternions are used to represent the rotations of the talking head. The subjective results show that our method can generate more naturalness talking head video than MakeItTalk. However, similar to MakeItTalk, our method also suffers from the pose limitation that the predicted poses are small. The main reason is that there is no explicit correlation between speech and the head pose, and the latter is more random. Therefore, using generative adversarial learning strategies can be considered in the future to predict random head movements that look natural. 

While the recently proposed method \cite{zhou2021pose} cannot be applied to an arbitrary person, this end-to-end method shows obvious superiority in generating more accurate lip movements compared to our landmark-based method. It is caused by the landmarks' low dimensionality that suppresses the details, which could lead to semantic mismatches between the landmarks and the photo-realistic face image. Some failure cases of generated images conditioned on landmarks are shown in Fig. \ref{fig:weak_landmark}. In this figure, the landmarks are extracted from the real target image, which means these landmarks are ground-truth landmarks in the talking head generation. However, even with these ground-truth landmarks, there are still differences between the generated images and real target images, as different lip shapes from the photo-realistic images could lead to the same sketch on a 2d plane due to reduced dimensions. However, the landmarks show the superiority on presenting the head pose, making the proposed method can predict the head pose automatically. For the future research, using head pose information provided by the landmarks can be considered in the end-to-end method to predict the head pose instead of using the head poses from a reference video as in \cite{zhou2021pose}.

\textbf{Ethical consideration}: While the proposed method can synthesize speech based on the input face image for any person from that input face image, we do not argue that there is an inevitable relation between a face and a voice. The proposed face-conditioned multi-speaker TTS module is not created to reconstruct someone's real voice but rather to give a face in a photo a voice that sounds as if the person in the photo could have produced speech with that voice. 
The proposed method could be used in many scenarios, e.g., film making, video editing, and human-computer interaction. However, such forward-looking technology could also have the potential to be misused or abused for various malevolent purposes, such as spreading false statements or misinformation. To prevent our released code from being abused, a watermark is included in this code to make the generated videos. We also encourage the public to report any suspicious videos to the appropriate authorities.

\section{Conclusion}
\label{sc:conclusion}

This paper presented a method, which we called AnyoneNet, which, for the first time, can synthesize a talking head video with synchronized speech for an arbitrary person with only text and a face image as input. The voice of the talking head is also created on the basis of the face image. The proposed method consists of two main modules, i.e., a face-conditioned multi-speaker TTS module and an audio-driven talking head video generation module. The results of several experiments showed that the proposed face-conditioned multi-speaker TTS can synthesize reasonable voices in harmony with the face in the given face image, and the proposed audio-driven talking head video generation method has state-of-the-art performance on the task of talking head generation.

\section*{Acknowledgment}
The authors thank Dong Wang et al. who built the CN-Celeb database provided us the real speaker identities in this database.

\bibliographystyle{IEEEtran}
\bibliography{Main.bbl}
%




\end{document}